\documentclass[conference]{IEEEtran}

\IEEEoverridecommandlockouts
\usepackage{array}
\usepackage{cite}
\usepackage{hyperref}
\usepackage{amsmath,amssymb,amsfonts}
\usepackage{algorithmic}
\usepackage{graphicx}
\usepackage{textcomp}
\usepackage{xcolor}
\usepackage{subfiles}
\usepackage[caption=false,font=footnotesize]{subfig}
\usepackage{microtype}
\usepackage{multirow}
\usepackage{makecell}
\usepackage{booktabs}
\usepackage{arydshln}
\def\BibTeX{{\rm B\kern-.05em{\sc i\kern-.025em b}\kern-.08em
    T\kern-.1667em\lower.7ex\hbox{E}\kern-.125emX}}
\begin{document}

\title{IMD: A 6-DoF Pose Estimation Benchmark for Industrial Metallic Objects\\
\thanks{*This work was conducted at ABB Corporate Research in V\"aster\aa s, Sweden, as part of a Master's thesis project.}
}

\author{\IEEEauthorblockN{Ruimin Ma*}
\IEEEauthorblockA{\textit{KTH Royal Institute of Technology}\\ 
Stockholm, Sweden \\
ruiminm@kth.se}
\and
\IEEEauthorblockN{Sebasti\'an Zudaire}
\IEEEauthorblockA{\textit{ABB Corporate Research}\\
Västerås, Sweden \\
sebastian.zudaire@se.abb.com}
\and
\IEEEauthorblockN{Zhen Li}
\IEEEauthorblockA{\textit{ABB Corporate Research}\\
Västerås, Sweden \\
zhen.li@se.abb.com}
\and
\IEEEauthorblockN{Chi Zhang}
\IEEEauthorblockA{\textit{ABB Corporate Research}\\
Västerås, Sweden \\
chi.zhang@se.abb.com}
}
\maketitle

\begin{abstract}
Object 6DoF (6D) pose estimation is essential for robotic perception, especially in industrial settings. It enables robots to interact with the environment and manipulate objects. However, existing benchmarks on object 6D pose estimation primarily use everyday objects with rich textures and low-reflectivity, limiting model generalization to industrial scenarios where objects are often metallic, texture-less, and highly reflective. To address this gap, we propose a novel dataset and benchmark namely \textit{Industrial Metallic Dataset (IMD)}, tailored for industrial applications. Our dataset comprises 45 true-to-scale industrial components, captured with an RGB-D camera under natural indoor lighting and varied object arrangements to replicate real-world conditions. The benchmark supports three tasks, including video object segmentation, 6D pose tracking, and one-shot 6D pose estimation. We evaluate existing state-of-the-art models, including XMem and SAM2 for segmentation, and BundleTrack and BundleSDF for pose estimation, to assess model performance in industrial contexts. Evaluation results show that our industrial dataset is more challenging than existing household object datasets. This benchmark provides the baseline for developing and comparing segmentation and pose estimation algorithms that better generalize to industrial robotics scenarios.
\end{abstract}


\section{Introduction}

Object 6-DoF (6D) pose estimation is essential for robotic perception. It plays a critical role in enabling autonomous robots to interact with the environment, particularly in industrial applications such as object manipulation, bin picking, and machine tending. Although recent deep learning and visual tracking methods have advanced performance on everyday-object benchmarks such as LINEMOD~\cite{linemod}, YCB-video~\cite{posecnn}, NOCS~\cite{NOCS}, these datasets mainly consist of textured, low-reflectivity household objects. Consequently, models trained and evaluated on them often fail to generalize to industrial environments, where objects tend to be metallic, texture-less, and highly reflective.

Industrial scenarios introduce further challenges for object detection and 6D pose estimation that everyday-object datasets do not capture, such as strong reflections, little surface texture, occlusions, and complex layouts.
Existing industrial collections such as T-LESS~\cite{T-Less} and MVTec ITODD~\cite{MVTEC} attempted to address this, but they remain limited in lighting diversity, material variety, and representation of true-scale metallic parts.

To bridge this gap, we present a new dataset, namely \textbf{Industrial Metallic Dataset (IMD)}, focused on industrial 6D pose estimation. It comprises 55 scenarios involving 45 real-world metallic components, each paired with a true-scale CAD model. Data were collected using an RGB-D camera mounted on a robot arm. Scenes were recorded in natural indoor lighting with varied object arrangements, including single parts, mixed sets, shape-similar groupings, and full object sets. Each scenario was captured under two camera configurations: top-down and 45-degree angled views, with about 200 frames per view, resulting in 110 video sequences and 256 annotated object sequences.

We benchmark state-of-the-art models on three tasks, including video object segmentation, 6D pose tracking, and one-shot 6D pose estimation. Our evaluation includes (1) cross-model comparison to identify model effectiveness, and (2) cross-dataset analysis to evaluate model performance in industrial domains. These results offer insights into the limitations of current approaches and highlight directions for improving industrial robotic vision.

The main contributions of this paper are:
\begin{itemize}
    \item \textbf{A purpose-built dataset of industrial metallic objects:} We introduce the IMD dataset designed for industrial applications, comprising 45 real-world metallic objects, each paired with a true-scale CAD model. The dataset includes 55 scenarios, with 110 video sequences and 256 annotated object sequences, recorded under natural indoor lighting using an Intel RealSense D405 RGB-D camera mounted on an ABB GoFa CRB 15000 robot, covering diverse object arrangements and realistic viewpoints.

    \item \textbf{High-quality 6D pose and segmentation annotations:} We provide precise ground-truth annotations for both segmentation and 6D object pose. Initial segmentation masks are generated using the Segment Anything Model 2 (SAM2)~\cite{sam2}, while 6D poses are computed via accurate coordinate-frame transformations from the robot to the camera space. The annotations are then manually refined to ensure pixel-level alignment with the CAD model silhouettes, resulting in high-fidelity labels suitable for training and evaluation.

    \item \textbf{Comprehensive benchmark evaluation and analysis across three tasks:} We benchmark video object segmentation using SAM2~\cite{sam2} and XMem~\cite{xmem}, 6D pose tracking using BundleTrack~\cite{bundletrack} and BundleSDF~\cite{bundleSDF}, and one-shot 6D pose estimation adapted from pose tracking models~\cite{oneshot}. Our analysis strategy includes: (1) cross-model comparison to identify algorithms best suited for the challenges of metallic, texture-less industrial components, and (2) cross-dataset evaluation to assess generalization from everyday-object datasets to industrial settings. Results show that our proposed IMD dataset is more challenging than existing everyday-object datasets.
\end{itemize}

\section{Related work}

\subsection{Datasets}
Existing benchmarks such as LINEMOD, YCB-video, and NOCS~\cite{linemod,posecnn,NOCS} have advanced 6D object pose estimation, but they primarily focused on everyday objects captured in controlled or domestic settings. For example, YCB-video dataset~\cite{posecnn} includes richly textured household items suitable for manipulation tasks, whereas LINEMOD~\cite{linemod} focuses on texture-less objects in cluttered but static scenes. These datasets commonly include items with simple geometries, saturated colors, and low-reflectivity materials. In contrast, industrial objects often exhibit metallic, texture-less, and highly reflective surfaces with severe specular highlights and complex lighting interactions. As a result, models perform well on everyday-object datasets may typically struggle to generalize to industrial scenarios, where material properties, lighting conditions, and environmental complexity introduce unique challenges.

Several datasets have been proposed to better simulate industrial scenarios to bridge this gap. T-LESS~\cite{T-Less} introduces 30 texture-less, symmetric objects, which are common in logistics and assembly lines. They were captured under varying viewpoints and lighting. MVTec ITODD~\cite{MVTEC} extends this by including surface defect annotations on industrial parts, focusing on real-world quality-inspection scenarios. BOP-Distribution~\cite{BOP-D} refines the benchmark evaluation through image-specific ambiguity labels that account for symmetries and occlusions, helping to make fairer comparisons between methods. Despite these advances, they still mainly focus on objects with matte, homogeneous surfaces or synthetic renderings, lacking the high reflectivity and intricate surface variations of true metallic parts in factory environments.

To address this gap, we propose the IMD dataset that includes 45 real-world metallic industrial objects with highly reflective surfaces. By varying object arrangements and capturing multi-view video sequences, we replicate practical industrial scenarios compared to prior datasets.

\subsection{Algorithms}

\subsubsection{Video object segmentation}

Video object segmentation (VOS) is an important component of pose estimation in many algorithms such as BundleTrack~\cite{bundletrack} and BundleSDF~\cite{bundleSDF}.
XMem~\cite{xmem} is designed for fully automatic, semi-supervised long-term VOS, propagating an initial mask annotation throughout the video without further user input. It achieves state-of-the-art results on long-video benchmarks with low GPU memory usage, enabling real-time performance on resource-constrained devices. More recent work SAM2~\cite{sam2} treats each frame independently while integrating temporal information via a learnable memory bank and cross-frame attention.
SAM2 was trained on the large and diverse Segment Anything Video (SA-V) dataset~\cite{sam2}, which is much larger than previous open-source datasets in scale and variety, such as DAVIS~\cite{DAVIS}, MOSE~\cite{MOSE}, and YouTubeVOS~\cite{youtube}, which were used to train XMem, and has shown strong performance in object video segmentation.

\subsubsection{Object 6D pose tracking}
Recent studies in object 6D pose estimation~\cite{review of pose estimation} can be grouped into three categories: CAD model-based methods on known objects, model-free tracking for novel objects, and hybrid methods that aim for cross-scenario generalization.
Early works such as PoseCNN~\cite{posecnn} leverage RGB or RGB-D images to estimate 6D poses via explicit decoupling of translation and rotation, and are especially effective in static or structured environments where object identities are known and annotated 3D models are available.
To address more open-set scenarios, model-free tracking methods such as BundleTrack~\cite{bundletrack} and BundleSDF~\cite{bundleSDF} eliminate the need for CAD models. 
BundleTrack~\cite{bundletrack} focuses on temporal coherence by maintaining a memory-augmented pose graph with VOS-guided segmentation propagation, whereas BundleSDF~\cite{bundleSDF} simultaneously reconstructs object geometry through neural fields and uses joint optimization for pose and shape. 
Recent unified approaches, such as FoundationPose~\cite{foudationpose}, push the boundary further by supporting both CAD-based and CAD-free modes within a single framework. They leverage large language model and diffusion model-aided synthetic data generation, neural implicit fields for novel view synthesis, and transformer-based hypothesis refinement to enable zero-shot generalization to unseen categories and instance-level tracking.
While the aforementioned methods show strong performance on everyday-object datasets, their effectiveness in industrial settings remains underexplored. In this paper, we investigate their performance in such environments.

\subsubsection{One-shot 6D pose estimation}
In many real-world scenarios, the target object may be previously unseen, with only a single observation available at test time without requiring instance-specific training, optionally accompanied by a CAD model. Sun et al.~\cite{onepose} formalized this setting as the ``one-shot pose estimation'' problem.
OnePose~\cite{onepose} and its extension OnePose++~\cite{onepose++} have explored this direction. OnePose~\cite{onepose} uses keypoint-based graph attention networks to aggregate consistent 3D features from multi-view correspondences, while~\cite{onepose++} builds upon detector-free feature matching~\cite{LOFTR} to reconstruct more complete models of low-texture objects, a common challenge in real-world environments.
Recent work PoseMatcher~\cite{posematcher} is designed to further improve the efficiency and robustness of template-based matching in one-shot scenarios. By employing a three-view training pipeline, it enables learning from scratch without relying on pre-trained descriptors.
Zudaire and Amadio~\cite{oneshot} proposed a one-shot 6D pose estimation approach, built upon existing pose tracking algorithms that support re-localization of the object after losing track of it. In this setting, pose tracking algorithms is first initialized by first processing a video tracking sequence for an object. Once initialized, individual frames, without time or spacial continuity to the video tracking sequence, can be processed independently to estimate the object's 6D pose. In this paper, we use this adjustment for both BundleTrack and BundleSDF for one-shot pose estimation.

\section{Methodology}
\subsection{Data capture}
\subsubsection{Objects set}
\begin{figure}
    \begin{center}
       \includegraphics[trim=400 50 100 50, clip, width=1\linewidth]{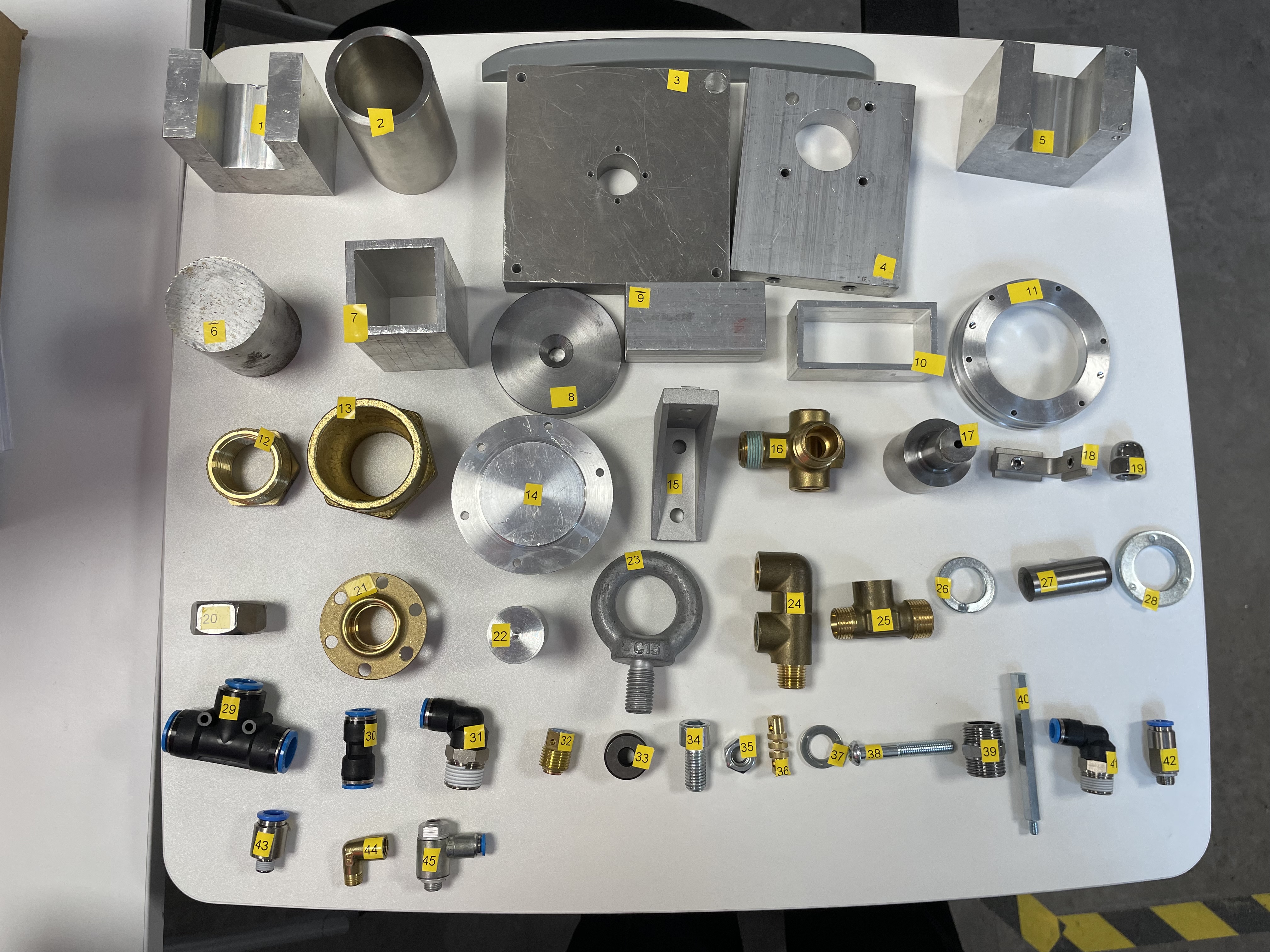} 
    \end{center}
    \caption{Overview of the objects used for data collection. Each object is labeled with a yellow ID tag for reference in this figure; note that these labels were removed during actual data collection.}
    \label{fig:Objects}
\end{figure}

As shown in Fig. \ref{fig:Objects}, a total of 45 industrial objects were selected with diameters ranging from 1.94~cm to 13.2~cm, a mean of 5.71~cm, and a standard deviation of 2.62~cm. The selection covers a wide range of materials and geometries typical of industrial applications, and reflects the diversity and physical constraints within the reachable workspace of robotic end-effectors in automated industrial systems. Each object is paired with a CAD model, which can be used for labeling or generating synthetic data.

This dataset intentionally focuses on metallic objects typical of industrial machine-tending scenario. Non-metallic categories (e.g., plastics, rubber, composites) are out of scope by design, since our goal is to capture perception failure modes unique to highly reflective, low-texture metals. Industrial dataset such as T-Less~\cite{T-Less} provide coverage for objects which are non-metallic, matte, texture-less.

\subsubsection{Dataset scenarios}
\begin{figure}
    \centering
    \begin{minipage}{0.48\linewidth}
        \centering
        \includegraphics[angle=90, width=\linewidth]{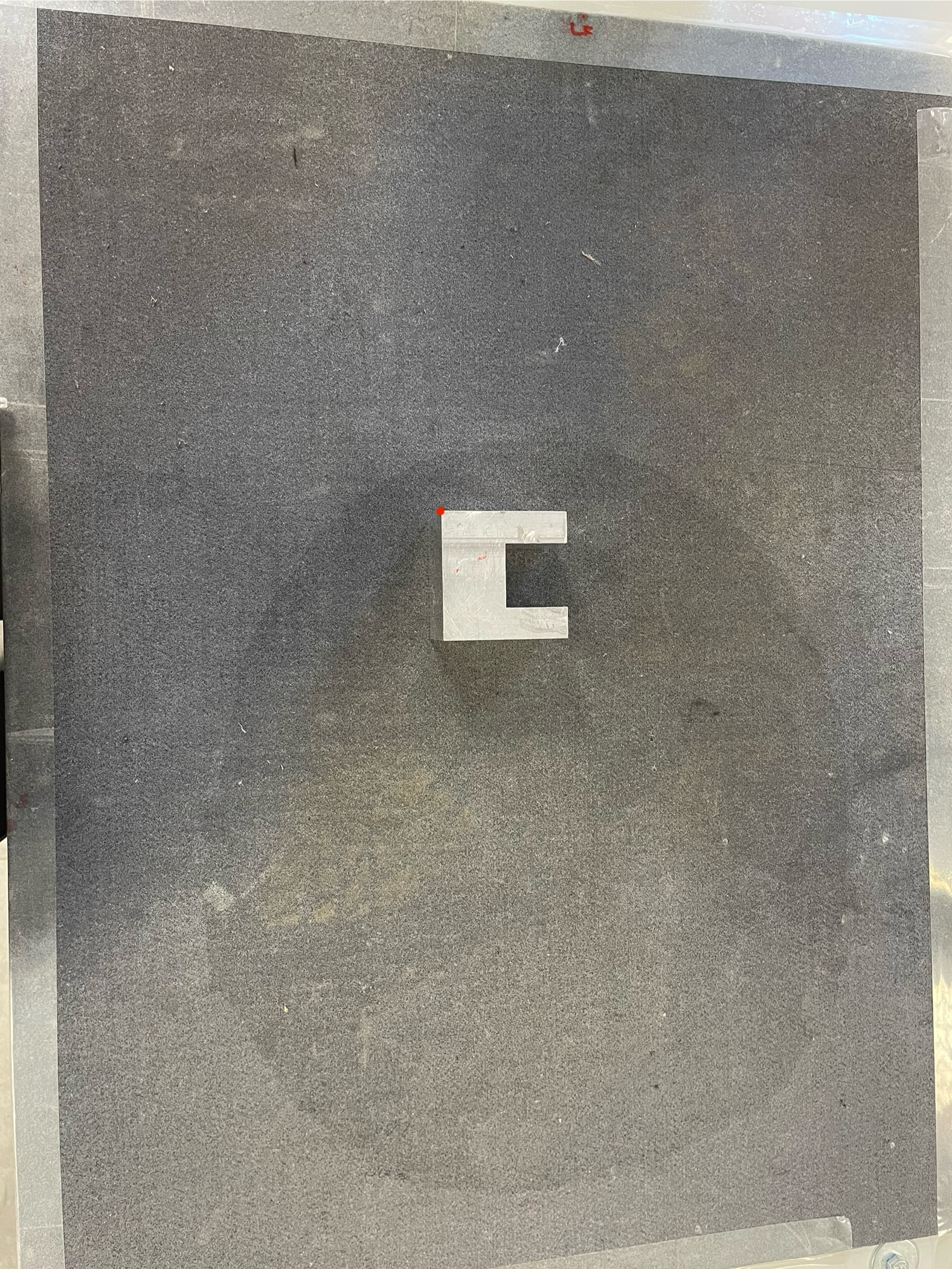}
        \footnotesize Single object
        \label{fig:single}
    \end{minipage}%
    \hfill
    \begin{minipage}{0.48\linewidth}
        \centering
        \includegraphics[width=\linewidth]{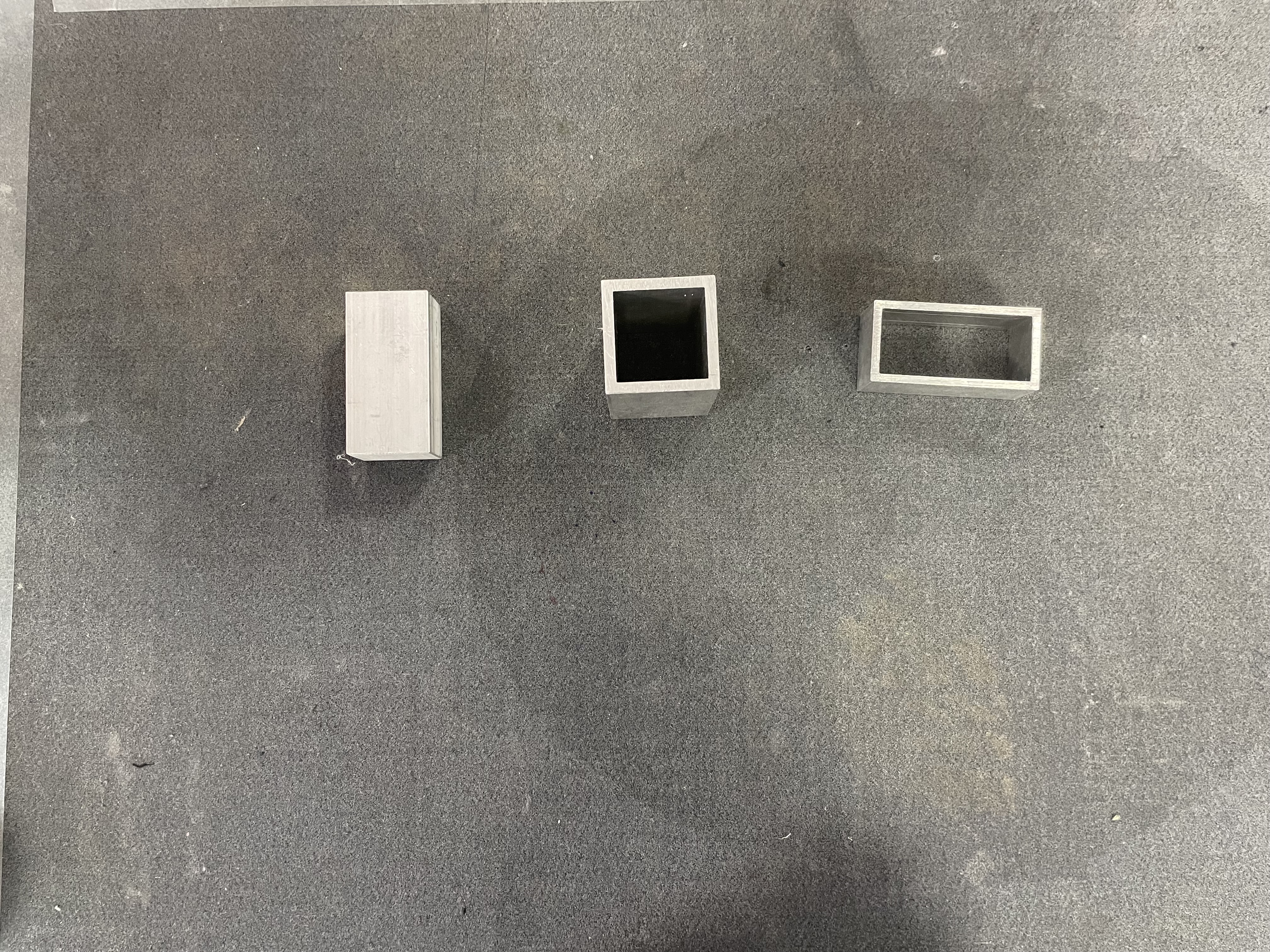}
        \footnotesize Similar shape objects
        \label{fig:similar}
    \end{minipage}
    \hfill
    \begin{minipage}{0.48\linewidth}
        \centering
        \includegraphics[width=\linewidth]{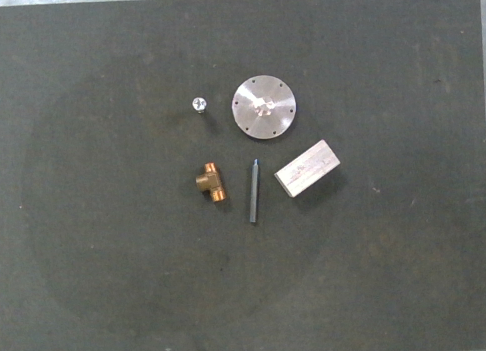}
        \footnotesize Random objects
        \label{fig:random}
    \end{minipage}
    \hfill
    \begin{minipage}{0.48\linewidth}
        \centering
        \includegraphics[angle=90, width=\linewidth]{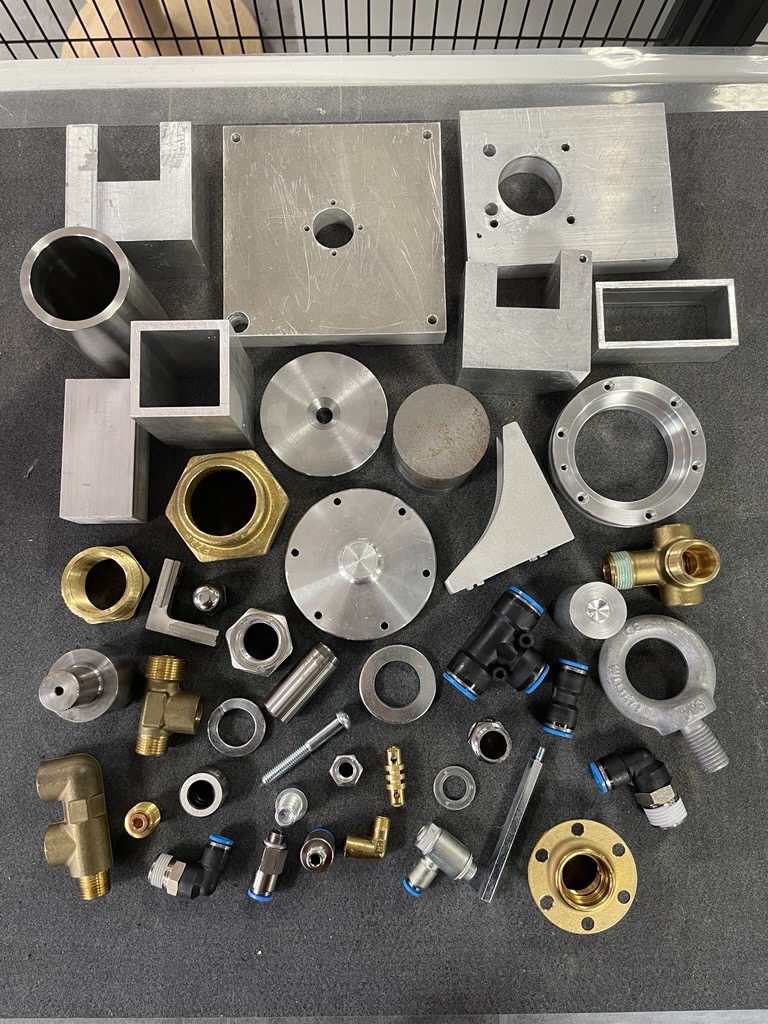}
        \footnotesize All objects
        \label{fig:all}
    \end{minipage}
    \caption{Examples of object arrangement scenarios used during data collection.}
    \label{fig:object_setup}
\end{figure}

To vary visual complexity and occlusion, the 45 objects were arranged in four scenario configurations: (1) single object, (2) shape-based groups of similar objects, (3) random groups of about five objects, and (4) all objects cluttered, as shown in Fig.~\ref{fig:object_setup}. All objects were placed on a matte-gray tabletop surface to replicate the material commonly used in conveyor belts. Only natural indoor daylight was used to capture realistic reflections and shadows.

\subsubsection{Capture frames}
We used an Intel RealSense D405 RGB-D camera (stereo depth, resolution $1280 \times 720$, field of view $87^\circ \,(H) \times 58^\circ \,(V)$, operating range 7\text{--}50 cm) mounted on the tool flange of a ABB GoFa CRB 15000 robot arm. We include two camera mounting configurations to capture different views: top-down view and 45-degree angled view. For the top-down view setup, the camera followed a square path above the working area, capturing 200 frames at 0.03 s intervals. For the 45-degree angled view, the camera followed a circular, 45-degree-inclined surrounding trajectory facing the working area, collecting 200 frames at 0.05 s intervals. We recorded RGB images, aligned with depth maps, and 6-DoF camera poses at each frame.
During capture, objects remained static while only the camera moved. We used a separate tool with a calibrated tool center point (TCP) as a reference to precisely locate each object within the world coordinate frame.
Across 45 objects, we set 55 scenarios and recorded 110 videos comprising 256 individual object sequences.

\subsection{Data annotation}
\subsubsection{Segmentation annotation}
We first generate initial segmentation masks using SAM2~\cite{sam2}, and then manually refine the mask with a multi-stage pipeline. We load the object's CAD model and apply a fixed rotation to align with the object's ground truth orientation. Using the recorded camera-world and object-world transforms, we compute the camera-object transform and convert its rotation into a Rodrigues vector. We then project each triangular face onto the image plane and rasterize those fully within the frame to produce a high-fidelity silhouette, unaffected by lighting or reflectance. Next, we interactively align this silhouette in the first frame by overlaying it on the RGB image and adjusting a horizontal offset $d_x$, vertical offset $d_y$, and rotation $\theta$ until it matches the pre-labeled mask. These offsets are then applied to every subsequent frame to correct systematic misalignments. To further tighten pixel-level correspondence, we compute the centroids of both the adjusted silhouette and the pre-labeled mask via image moments for each frame and translate the silhouette so its centroid coincides with that of the pre-labeled mask. Finally, we manually select the more accurate mask for each frame, retaining SAM2’s output when it handles occlusion better, or the rendered silhouette when it captures true contours more precisely.

\subsubsection{6D pose annotation}
To obtain the ground-truth object 6D pose in each camera frame, we first pre-label the pose using pose transformations, followed by manual refinement. 
For each predefined camera frame \(i\), the camera pose $\mathbf{T}_{w_i}^c$ with respect to the world coordinate is represented as:
\begin{equation}
    {T}_{w_i}^c = \begin{bmatrix} {R}_{wc_i} & {t}_{wc_i} \\ {0} & 1 \end{bmatrix} \quad
    \label{eq:1}
\end{equation}
\noindent where $\mathbf{R}_{wc_i}$ and $\mathbf{t}_{wc_i}$ are the rotation $3\times3$ and translation $3\times1$ components, respectively.

Using similar notation, the object pose $\mathbf{T}_w^o$ in the world coordinate remains fixed during each data capture sequence and is defined as below, which is recorded using a calibrated TCP:
 \begin{equation}
    \quad {T}_w^o = \begin{bmatrix} {R}_{wo} & {t}_{wo} \\ {0} & 1 \end{bmatrix}
    \label{eq:2}.
\end{equation}
The ground-truth object pose $T_c^o$ with respect to each camera frame \(i\) is then calculated by:
\begin{align}
    T_c^o &= (T_w^c)^{-1} T_w^o
    \label{eq:3}
\end{align}

In this way, we obtained the initial annotation. The pose is also manually refined to avoid misalignment.

\begin{figure}
    \centering
    \begin{minipage}{0.48\linewidth}
        \centering
        \includegraphics[trim=0 30 0 30, clip, width=\linewidth]{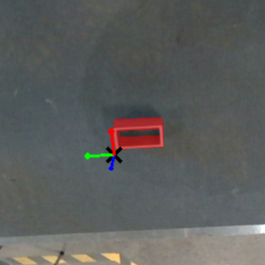}
        \footnotesize (a) Frame 0
        \vspace{0.5em}
        \label{fig:fram0}
    \end{minipage}%
    \hfill
    \begin{minipage}{0.48\linewidth}
        \centering
        \includegraphics[trim=0 30 0 30, clip, width=\linewidth]{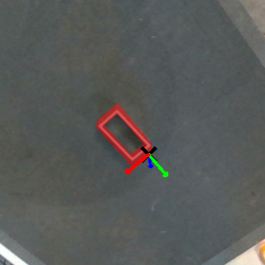}
        \footnotesize (b) Frame 50
        \vspace{0.5em}
        \label{fig:frame50}
    \end{minipage}
    \hfill
    \begin{minipage}{0.48\linewidth}
        \centering
        \includegraphics[trim=0 30 0 30, clip, width=\linewidth]{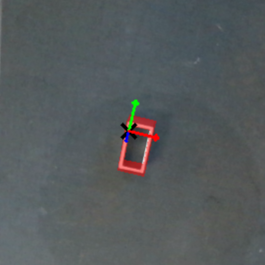}
        \footnotesize (c) Frame 100
        \label{fig:frame100}
    \end{minipage}
    \hfill
    \begin{minipage}{0.48\linewidth}
        \centering
        \includegraphics[trim=0 30 0 30, clip, width=\linewidth]{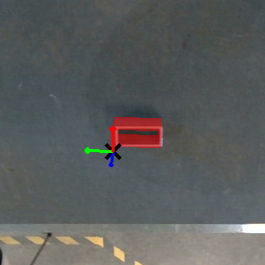}
        \footnotesize (d) Frame 150
        \label{fig:frame150}
    \end{minipage}
    \caption{Visualization of the ground-truth 6D pose of the metallic hollow cubic object at four time steps. Each image shows the segmentation mask and the projected ground-truth pose axes (x: red, y: green, z: blue) at frames 0, 50, 100, and 150.}
    \label{fig:visualization}
\end{figure}

An example of pose and segmentation annotation under the surround-view setup is shown in Fig. \ref{fig:visualization}.

\subsection{Experiments and prediction baselines}
\subsubsection{Cross-model evaluation}
We evaluate model performance on three tasks: segmentation, 6D pose tracking, and one-shot 6D pose estimation. For all algorithms we use their pretrained, open-source models without training or finetuning. We implement, evaluate, and compare the following state-of-the-art models:

\textbf{Object video segmentation:}

\begin{itemize}
    \item {XMem}: Proposed by Cheng et al. in 2022, XMem is a video object segmentation model that achieves accurate temporal consistency across frames while maintaining low memory usage through external memory mechanisms.
    \item {SAM2}: Proposed by Ravi et al. in 2024, SAM2 is a unified, promptable model for both image and video segmentation. It enables efficient, real-time segmentation and tracking of arbitrary objects based on user-provided prompts.
\end{itemize}
We provide the ground-truth mask from the first frame as input. XMem and SAM2 then predict segmentation masks for the remaining frames in the video sequence.

\textbf{6D pose tracking:}

\begin{itemize}
    \item {BundleTrack}: Proposed by Wen et al. in 2021, BundleTrack is a real-time 6D pose tracking framework that combines deep learning-based segmentation and feature extraction with memory-augmented pose graph optimization to track unknown rigid objects in video sequences.
    \item {BundleSDF}: Proposed by Wen et al. in 2023 as an extension of BundleTrack, BundleSDF performs near real-time 6-DoF pose tracking and 3D reconstruction of unknown rigid objects using signed distance field optimization and dense visual alignment.
\end{itemize}

Models are given the ground-truth segmentation mask and pose of the first frame. They are then required to estimate the object poses across the subsequent frames.

\textbf{One-shot 6D pose estimation}

We use BundleTrack and BundleSDF under the one-shot pose estimation setting, as proposed in~\cite{oneshot}. Given a recorded video (i.e., collection of frames) for a specific object, the evaluation procedure is as follows:

\begin{itemize}
    \item BundleTrack and BundleSDF are initialized by tracking the object over the first 50\% of the video frames. The inputs for this tracking process are the same as for 6D pose tracking.
    \item The remaining 50\% of the frames will be used to feed one at a time to BundleTrack and BundleSDF. Models are configured to ensure none of the frames from the remaining 50\% are added to the memory pool or used for pose estimation, thus breaking temporal and spacial coherence after the first frames of the remaining 50\%, ensuring a one-shot pose estimation setting.
\end{itemize}

\subsubsection{Cross-dataset evaluation}
In addition to cross-model evaluation, we also perform cross-dataset evaluation to compare the characteristics of the proposed IMD dataset with existing everyday-object benchmarks. The following datasets are used for comparison:

\begin{itemize}
    \item {Segmentation:} DAVIS-2017~\cite{DAVIS} dataset is used to evaluate segmentation performance on video sequences with everyday objects.
    \item {6D Pose Estimation:} YCB-video dataset~\cite{posecnn} is used to benchmark pose estimation in non-industrial settings. we use a representative subset from sequence 0048 to 0059 for testing following the existing studies like~\cite{Densefusion}. The 12 video sequences contains 55 object sequences. In our settings, we down-sampled the video sequence into 6~fps.
\end{itemize}

\subsection{Evaluation methods}
\subsubsection{Segmentation}

Performance is quantitatively evaluated using the Intersection over Union (IoU) metric as shown in Eq.~\ref{eq:iou}, also known as Jaccard Index, which measures the overlap between ground-truth annotations $B_{gt}$ and model predictions $B_p$ at the pixel level.
\begin{equation}
    \text{IoU} = \frac{|B_p \cap B_{gt}|}{|B_p \cup B_{gt}|}
    \label{eq:iou}
\end{equation}

\subsubsection{Pose estimation evaluation}
For 6D pose tracking and one-shot pose estimation, we evaluate translation and rotation errors, which are typically used for robotic object manipulation. These metrics disentangle the errors into two distinct sources, enabling more precise analysis and supporting the development of more effective grasping strategies.
\begin{itemize}
    \item {Translation error (TE)}, computed as the Euclidean distance between the predicted and ground-truth object centroids in 3D space.
    \item {Rotation error (RE)}, defined as the angular deviation between the predicted and ground-truth rotation matrices, measured in degrees.
\end{itemize}

Both metrics are computed per frame for each object instance and subsequently averaged to obtain per object sequence and overall performance scores. 

\section{Results and discussions}

\subsection{Video object segmentation}

We evaluate the XMem~\cite{xmem} and SAM2~\cite{sam2} models on both the proposed IMD dataset and the DAVIS-2017 dataset~\cite{DAVIS}. Quantitative results using IOU (also known as Jaccard Index) are presented in Table~\ref{tab:seg_result}. 

\begin{table}[h]
    \caption{Quantitative comparisons of segmentation performance IoU on DAVIS-2017 and the proposed IMD dataset. ``Recall'' here refers to recall at an IOU threshold of 0.5. Higher values indicate better performance.}
    
    \label{tab:seg_result}
    \centering
        \begin{tabular}{lrrrr}
        \toprule
        Method & \multicolumn{2}{c}{DAVIS-2017} & \multicolumn{2}{c}{IMD (Ours)} \\
        \cmidrule(lr){2-3} \cmidrule(lr){4-5}
        & Mean & Recall & Mean & Recall \\
        \midrule
        XMem & 0.863 & 1.000 & 0.746 & 0.922 \\
        SAM2 & \textbf{0.893} & 1.000 & \textbf{0.770} & \textbf{0.980} \\
        \bottomrule
    \end{tabular}
\end{table}

\begin{figure}[t]
    \centering
\includegraphics[width=1\linewidth]{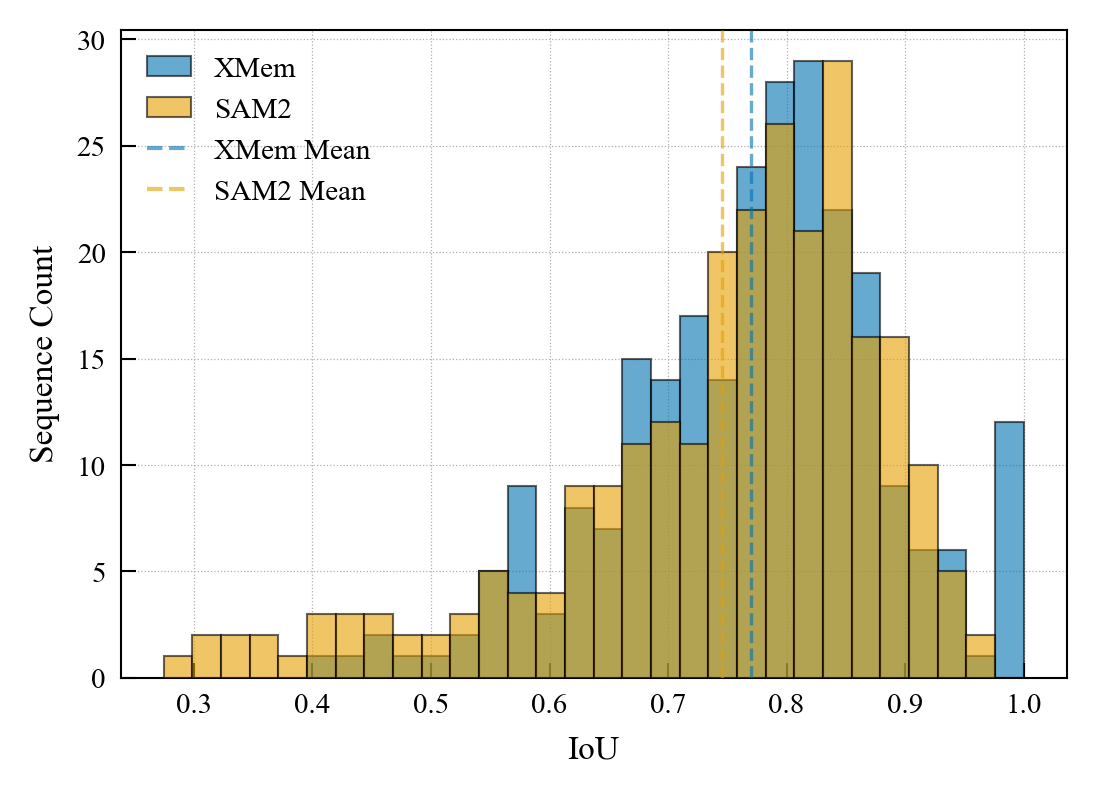}
    \caption{Distribution of sequence-level segmentation IoU scores for XMem (blue) and SAM2 (yellow) on the proposed IMD dataset. The bars represent the number of sequences at each average IoU value, and the dashed vertical lines indicate the mean IoU for each method.}
    \label{fig:mex_sam2_xmem}
\end{figure}

Cross-model comparison shows that SAM2 consistently outperforms XMem on both datasets. On DAVIS-2017 dataset, SAM2 achieves a higher mean IOU of 0.893 compared to 0.863 for XMem, indicating more accurate and consistent segmentation. On the IMD dataset, SAM2 achieves a slightly higher mean IOU of 0.770 compared to 0.746 for XMem, and a higher recall of 0.980 at an IoU threshold of 0.5 compared to 0.922 for XMem, demonstrating better robustness under challenging industrial conditions. Fig.~\ref{fig:mex_sam2_xmem} compares the IoU distribution of both models on the IMD dataset. SAM2’s scores are concentrated between the 25th percentile at 0.700 and the 75th percentile at 0.852, closely aligned with XMem’s range of 0.679 to 0.851.

While SAM2 shows better segmentation results on both everyday and industrial objects, it demands greater memory and longer inference time compared to XMem. Therefore, the choice between these methods should be guided by the specific requirements of the use case.

Cross-dataset evaluation shows that both models encounter challenges on the IMD dataset compared to the DAVIS-2017 benchmark due to the reflective and texture-less nature of industrial objects. On DAVIS-2017, XMem and SAM2 achieve high mean IoU scores of 0.863 and 0.893, respectively, with a recall of 1.0 at an IoU threshold of 0.5, indicating accurate and consistent segmentation across sequences. However, performance degrades noticeably on the IMD dataset. XMem's mean IOU decreases by 0.117 to 0.746, and its recall drops to 0.922. SAM2's mean IOU decreases by 0.123 to 0.770 and its recall drops to 0.980. These results highlight the increased difficulty of industrial scenarios and the limitations of current segmentation models in handling such scenarios.

\begin{table}[htbp]
\caption{Quantitative comparison of average translation error (TE) and average rotation error (RE) on the YCB-Video dataset and the proposed IMD dataset under top-down and 45-degree angled views. Upper section: pose tracking results. Lower section: one-shot estimation results. Lower values indicate better performance. }
\label{tab:pose_tracking_comparison}
\centering
\begin{tabular}{l m{0.7cm} m{0.7cm} m{0.7cm} m{0.7cm} m{0.7cm} m{0.7cm}}
\toprule
Method
& \multicolumn{2}{c}{YCB-video}
& \multicolumn{2}{c}{IMD-top down} 
& \multicolumn{2}{c}{IMD-45\textdegree view} \\
\cmidrule(lr){2-3} \cmidrule(lr){4-5} \cmidrule(lr){6-7}
& TE (mm) & RE (\textdegree)
& TE (mm) & RE (\textdegree)
& TE (mm) & RE (\textdegree) \\
\midrule
BundleTrack & \textbf{2.26}  & \textbf{4.48}  & \textbf{6.61}  & \textbf{8.12}  & \textbf{32.23} & \textbf{49.17} \\
BundleSDF   & 5.64  & 8.09  & 8.82  & 13.08 & 32.95 & 58.27 \\
\hdashline
\makecell[l]{BundleTrack\\(One-shot)} & 119.55 & 52.39 & 95.09 & 7.00  & 65.60 & 54.48 \\
\makecell[l]{BundleSDF\\(One-shot)}   & \textbf{6.80}  & \textbf{17.61} & \textbf{10.46} & \textbf{17.98} & \textbf{42.44} & \textbf{72.21} \\
\bottomrule
\end{tabular}
\end{table}

\subsection{Pose tracking evaluation}

We evaluate BundleTrack~\cite{bundletrack} and BundleSDF~\cite{bundleSDF} on the YCB-video dataset~\cite{posecnn} and the proposed IMD dataset, under two camera configurations with the top-down and 45-degree angled views. Quantitative results based on translation error and rotation error are presented in Table~\ref{tab:pose_tracking_comparison}. BundleSDF fails to track objects when they disappear from the camera view due to occlusion or experience large viewpoint changes, and reports failure in such cases.
We included only the sequences successfully tracked by BundleSDF. For BundleTrack, when tracking or one-shot estimation fails, the pose is reset to the center of the image. Results are reported for 49 out of 55 object sequences from YCB-video, 117 out of 128 from IMD top-down view, and 106 out of 128 from IMD 45-degree view.

Cross-model evaluation shows that BundleTrack consistently outperforms BundleSDF across datasets for object 6D pose tracking. As shown in Table~\ref{tab:pose_tracking_comparison}, BundleTrack achieves a lower average translation error of 2.26~mm and rotation error of 4.48~degrees on the YCB-video dataset, compared to 5.64 mm and 8.09 degrees for BundleSDF. On the IMD top-down view, BundleTrack again performs better, with 6.61 mm and 8.12 degrees, while BundleSDF reaches 8.82 mm and 13.08 degrees. For IMD with 45-degree angled view, both methods show degraded performance, while BundleTrack continues to slightly outperform BundleSDF.

\begin{figure}[t]
    \centering
    \includegraphics[width=1\linewidth]{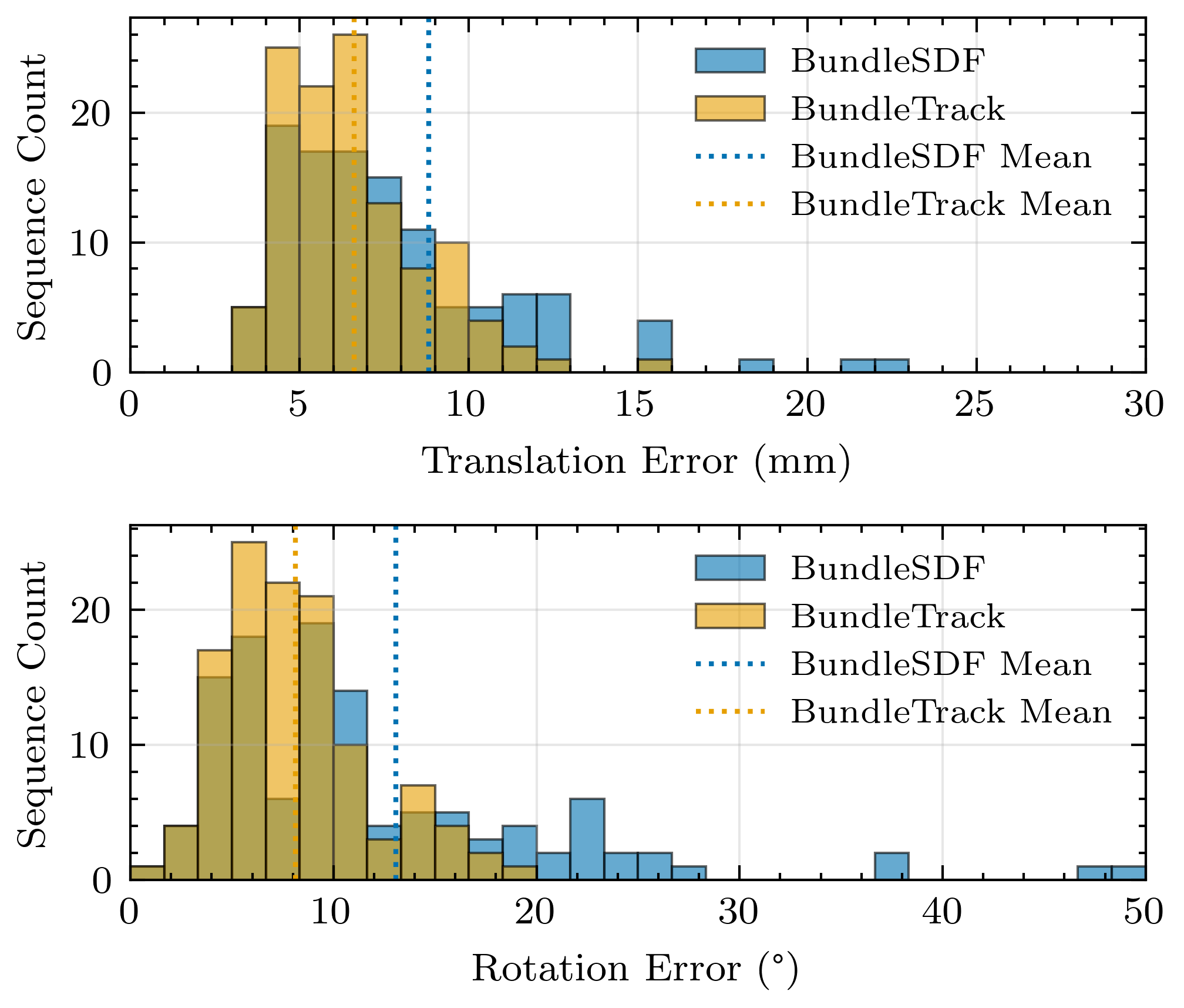}
    \caption{Distribution of sequence-level tracking error for BundleSDF (blue) and BundleTrack (yellow) on the proposed IMD dataset captured from the top-down view. Top: translation errors in mm. Bottom: rotation errors in degrees. Vertical dotted lines indicate the mean error for each method.}
    \label{fig:mex_tracking_2models_top}
\end{figure}

Fig.~\ref{fig:mex_tracking_2models_top} shows the error distributions for both models on the IMD dataset with top-down view. BundleTrack's histograms are more tightly concentrated around the median, whereas BundleSDF shows higher variance and longer tails in rotation error. For example, BundleTrack's translation error spans from 4.98 mm at the 25th percentile to 7.72 mm at the 75th percentile, with a median of 6.26 mm. In contrast, BundleSDF spans from 4.98 mm to 9.88 mm, with a median of 7.10 mm. These results indicate that for pure tracking tasks, BundleTrack delivers strong performance, outperforming BundleSDF in simpler scenarios and remaining competitive under more challenging conditions.

\begin{figure*}[htb]
    \centering
    \begin{minipage}{0.24\linewidth}
        \centering
        \includegraphics[trim=0 40 0 15, clip, width=\linewidth]{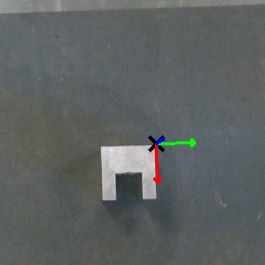}
        \label{fig:fram0_track}
    \end{minipage}
    \hfill
    \begin{minipage}{0.24\linewidth}
        \centering
        \includegraphics[trim=0 40 0 15, clip, width=\linewidth]{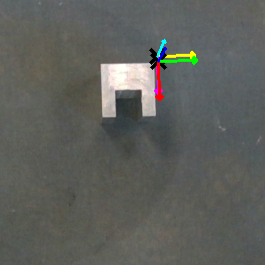}
        \label{fig:frame25_track}
    \end{minipage}
    \hfill
    \begin{minipage}{0.24\linewidth}
        \centering
        \includegraphics[trim=0 40 0 15, clip, width=\linewidth]{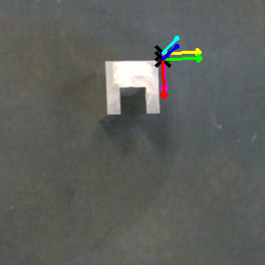}
        \label{fig:frame_50_track}
    \end{minipage}
    \hfill
    \begin{minipage}{0.24\linewidth}
        \centering
        \includegraphics[trim=0 40 0 15, clip, width=\linewidth]{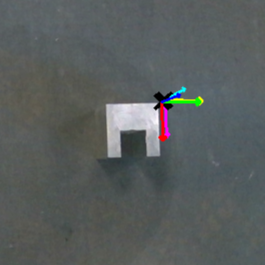}
        \label{fig:frame75_track}
    \end{minipage}
    \begin{minipage}{0.24\linewidth}
        \centering
        \includegraphics[trim=0 40 0 15, clip, width=\linewidth]{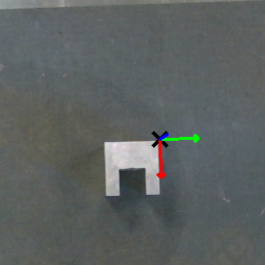}
        \footnotesize Frame 0
        \label{fig:fram0_sdf_top}
    \end{minipage}
    \hfill
    \begin{minipage}{0.24\linewidth}
        \centering
        \includegraphics[trim=0 40 0 15, clip, width=\linewidth]{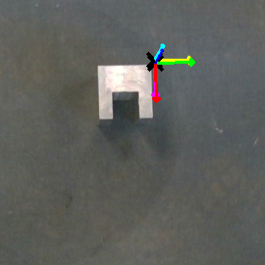}
        \footnotesize Frame 25
        \label{fig:frame25_sdf_top}
    \end{minipage}
    \hfill
    \begin{minipage}{0.24\linewidth}
        \centering
        \includegraphics[trim=0 40 0 15, clip, width=\linewidth]{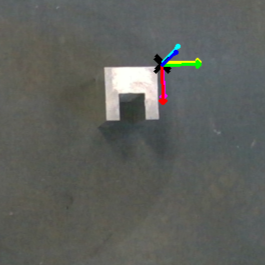}
        \footnotesize Frame 50
        \label{fig:frame_50_sdf_top}
    \end{minipage}
    \hfill
    \begin{minipage}{0.24\linewidth}
        \centering
        \includegraphics[trim=0 40 0 15, clip, width=\linewidth]{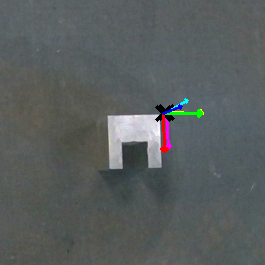}
        \footnotesize Frame 75
        \label{fig:frame75_sdf_top}
    \end{minipage}
    \caption{Example pose tracking results on the IMD dataset captured from the top-down view. The top row shows BundleTrack results, and the bottom row shows BundleSDF results. Frames 0, 25, 50, and 75 are displayed. Ground-truth 6-DoF axes are shown in red (x), green (y), and blue (z), while estimated pose axes are shown in cyan (x), yellow (y), and magenta (z).}
    \label{fig:top-visual}
\end{figure*}

\begin{figure*}[htb]
    \centering
    \begin{minipage}{0.24\linewidth}
        \centering
        \includegraphics[trim=0 30 0 30, clip, width=\linewidth]{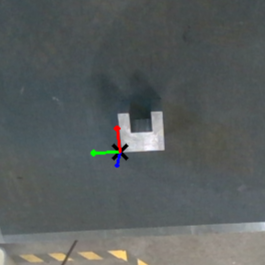}
        \label{fig:fram0_sdf}
    \end{minipage}
    \hfill
    \begin{minipage}{0.24\linewidth}
        \centering
        \includegraphics[trim=0 30 0 30, clip, width=\linewidth]{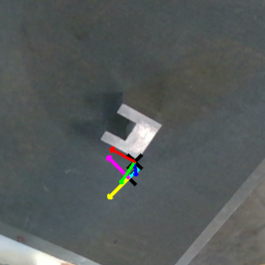}
        \label{fig:frame25_sdf}
    \end{minipage}
    \hfill
    \begin{minipage}{0.24\linewidth}
        \centering
        \includegraphics[trim=0 30 0 30, clip, width=\linewidth]{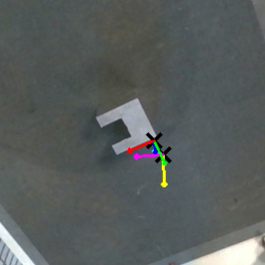}
        \label{fig:frame_50_sdf}
    \end{minipage}
    \hfill
    \begin{minipage}{0.24\linewidth}
        \centering
        \includegraphics[trim=0 30 0 30, clip, width=\linewidth]{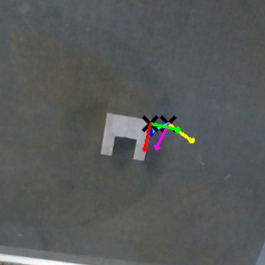}
        \label{fig:frame75_sdf}
    \end{minipage}
    \begin{minipage}{0.24\linewidth}
        \centering
        \includegraphics[trim=0 30 0 30, clip, width=\linewidth]{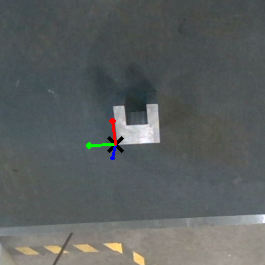}
        \footnotesize Frame 0
        \label{fig:fram0_sdf_surround}
    \end{minipage}
    \hfill
    \begin{minipage}{0.24\linewidth}
        \centering
        \includegraphics[trim=0 30 0 30, clip, width=\linewidth]{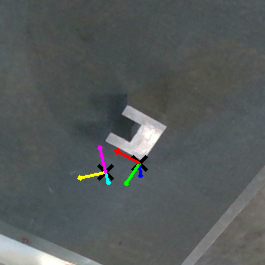}
        \footnotesize Frame 25
        \label{fig:frame25_sdf_surround}
    \end{minipage}
    \hfill
    \begin{minipage}{0.24\linewidth}
        \centering
        \includegraphics[trim=0 30 0 30, clip, width=\linewidth]{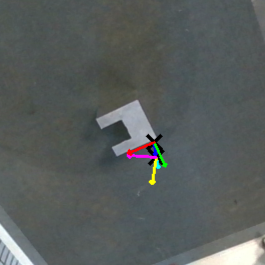}
        \footnotesize Frame 50
        \label{fig:frame_50_sdf_surround}
    \end{minipage}
    \hfill
    \begin{minipage}{0.24\linewidth}
        \centering
        \includegraphics[trim=0 30 0 30, clip, width=\linewidth]{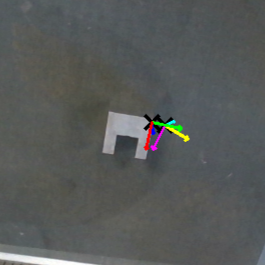}
        \footnotesize Frame 75
        \label{fig:frame75_sdf_surround}
    \end{minipage}
    \caption{ Example pose tracking results on the IMD dataset captured from the 45-degree angled view. The top row shows BundleTrack results, and the bottom row shows BundleSDF results. Frames 0, 25, 50, and 75 are displayed. Ground-truth 6-DoF axes are shown in red (x), green (y), and blue (z), while estimated pose axes are shown in cyan (x), yellow (y), and magenta (z). }
    \label{fig:surround_visual}
\end{figure*}

Cross-dataset comparison reveals that the performance of both BundleTrack and BundleSDF degrades on the proposed IMD dataset compared to the YCB-video dataset. For instance, BundleTrack's average translation increases from 2.26~mm on YCB-video to 6.61~mm on the IMD top-down view, and further deteriorate to 32.23~mm under the IMD 45-degree angled view.
Figs.~\ref{fig:top-visual} and ~\ref{fig:surround_visual} illustrate pose tracking results of BundleTrack and BundleSDF versus ground-truth at frames 0, 25, 50, and 70, for the top-down and 45-degree angled views, respectively. The comparison highlights the models’ sensitivity to industrial variability, especially under large viewpoint shifts in the 45-degree angled configuration. These results indicate that neither method generalizes reliably in complex industrial scenarios, highlighting the need for pose tracking approaches focusing on industrial environments. The proposed IMD dataset introduces additional challenges of feature matching across frames due to reflective and low-texture surfaces.

\subsection{One-shot 6D pose estimation evaluation}

We evaluate the one-shot 6D pose estimation setting~\cite{oneshot}for BundleTrack~\cite{bundletrack} and BundleSDF~\cite{bundleSDF} on the YCB-video dataset~\cite{posecnn} and the proposed IMD dataset under two camera configurations, top-down and 45-degree angled views. The bottom half of Table~\ref{tab:pose_tracking_comparison} reports translation and rotation errors.

Cross-model comparison shows that BundleSDF consistently outperforms BundleTrack across all datasets in the one-shot setting. BundleTrack fails to produce reasonable results and defaults to a fixed pose, whereas BundleSDF remains functional, demonstrating better robustness under single frame input.

Compared to pose tracking, one-shot estimation with BundleSDF shows degraded performance. For example, on the YCB-video dataset, translation error increases from 5.64~mm for tracking to 6.80~mm (+20.6\%) for one-shot, while the rotation error increases substantially from 8.09~degree to 17.61~degree (+117.7\%). Similarly, on the IMD top-down view data, translation error increases from 8.82~mm for tracking to 10.46~mm (+18.6\%) for one-shot, and rotation error from 13.08~degree to 17.98~degree (+37.5\%). The larger increase in rotation error suggests that orientation estimation is more challenging under one-shot settings.

Cross-dataset comparison shows that one-shot 6D pose estimation is challenging for industrial scenarios. For BundleSDF in the one-shot configuration, translation error decreases from 6.80~mm for YCB-video dataset to 10.46~mm for IMD top-down view data, highlighting the difficulty of generalizing to industrial objects with reflective and low-texture surfaces.

One possible reason for the degraded performance on the proposed IMD dataset is the reflective and texture-less surfaces of industrial objects, which negatively impacts two crucial components of both BundleTrack and BundleSDF. The first one is that the depth information coming from the RGB-D camera are often unreliable on such surfaces. Second, the feature detection and matching algorithms, Lf-Net~\cite{lfnet} in BundleTrack and LoFTR~\cite{LOFTR} in BundleSDF, will struggle to find meaningful matches between frames. These limitations hinder the ability of both methods to obtain robust pose estimation in industrial scenarios.

This can be further seen in the most challenging IMD 45-degree angled view data, were both pose tracking and one-shot estimation of BundleSDF yield large errors, making them impractical for real-world industrial applications. These findings highlight the need for more robust algorithms tailored to complex industrial scenarios with extreme viewpoint variations and texture-deficient surfaces.

In addition to cross-dataset analysis, it is informative to compare IMD dataset with existing industrial datasets. The BOP-Distrib benchmark~\cite{BOP-D} highlights object symmetries and partial occlusions, where methods show only minor losses on YCB-Video dataset but much larger degradations on T-LESS~\cite{T-Less} dataset due to its many texture-less, symmetric objects under occlusion. In contrast, the proposed IMD dataset emphasizes a different aspect of industrial difficulty: low-texture, highly reflective metallic parts. Notably, both datasets has shown that the industrial objects present greater challenges than typical household items.

\section{Conclusion}
In conclusion, we introduced the Industrial Metallic Dataset (IMD), a novel benchmark designed to evaluate object segmentation and 6D pose estimation for industrial objects. Unlike existing datasets that focus on everyday objects, IMD features 45 metallic, low-texture, and reflective components, closely reflecting real-world industrial conditions. Through evaluation of state-of-the-art models including XMem and SAM2 for segmentation, and BundleTrack and BundleSDF for pose estimation, we demonstrate that current methods struggle to generalize to these challenging industrial scenarios, especially under 45-degree angled view and in one-shot pose estimation settings. Our research fills the gap in current benchmarks and provides a baseline for advancing perception algorithms that are robust, accurate, and practical for real-world industrial robotics.

\section*{Acknowledgment}
We thank Daniel Andersson for providing the industrial objects, thank Jiangwei Huang for preparing the CAD models, and thank Dr. Liwei Qi for providing support for the whole project. We also thank Dr. Biao Zhang and Dr. Haoyan Liu for their valuable discussions during the development of the dataset.


\end{document}